%% file: template.tex
\documentclass[a4paper]{article}

\usepackage{INTERSPEECH2018}
\usepackage{pgfplots}
\usepackage{amsmath}
\usepackage[subtle]{savetrees}
\pgfplotsset{compat=1.14}
\usepgfplotslibrary{groupplots}
\DeclareMathOperator{\proj}{proj}
\title{State Gradients for RNN Memory Analysis}
\name{Lyan Verwimp, Hugo Van hamme, Vincent Renkens, Patrick Wambacq}
\address{
  ESAT -- PSI, KU Leuven, Belgium}
\email{\{lyan.verwimp, hugo.vanhamme, vincent.renkens, patrick.wambacq\}@esat.kuleuven.be}

\begin{document}

\maketitle
\begin{abstract}
We present a framework for analyzing what the state in RNNs remembers from its input embeddings. Our approach is inspired by backpropagation, in the sense that we compute the gradients of the states with respect to the input embeddings.~The gradient matrix is decomposed with Singular Value Decomposition to analyze which directions in the embedding space are best transferred to the hidden state space, characterized by the largest singular values. 
We apply our approach to LSTM language models and investigate to what extent and for how long certain classes of words are remembered on average for a certain corpus. Additionally, the extent to which a specific property or relationship is remembered by the RNN can be tracked by comparing a vector characterizing that property with the direction(s) in embedding space that are best preserved in hidden state space.
\end{abstract}

\section{Introduction}

Neural networks are remarkably powerful models that are currently the state of the art in many applications, such as speech recognition (e.g.~\cite{Han}), language modeling~\cite{Merity,Merity18} and image recognition (e.g.~\cite{Szegedy-inception}). However, one of the main disadvantages often mentioned is the fact that they remain `black-box' models, meaning that what the network has learned can only be explained in terms of its weights, that are not interpretable by humans. Very often, extensions of existing models are proposed based on how we \textit{think} that the models function. However, an alternative approach is first trying to better understand how and why neural networks work, and then, if new insights are gained, improving the models based on those insights.


We propose a framework to investigate what the states of recurrent neural network models (RNNs) remember from their input and for how long. We apply our approach to the current state of the art in language modeling, long short-term memory~\cite{Hoch} (LSTM) LMs~\cite{Sun:12}, but it can be applied to other types of RNNs too and to other models with continuous word representations as input. Usually, it is not straightforward to see how an input word embedding is encoded in the state of the neural network, because the latter is the result of a series of weight multiplications, nonlinearities and combinations with the states of the previous time steps. Our framework sheds more light on the relationship between the input and the state. 




Our framework is inspired by backpropagation, the algorithm that is used for training neural networks based on the gradient of the loss with respect to the weights. Instead of computing the gradient of the loss, we compute the gradient of the state with respect to the input embedding to capture the influence of the input on the state.
To examine how long input words are remembered by the RNN, we calculate the gradient with a certain delay -- with respect to the input word embedding a few time steps earlier, which is similar to the unfolding during backpropagation through time. 
The gradient matrix is decomposed with Singular Value Decomposition (SVD) and the relationship between the singular values (SVs), that indicate how much of the input embeddings is preserved in the state, and the delay is inspected. This relationship can be investigated for a gradient matrix averaged over all words, over specific classes of words or over occurrences of individual words.

Additionally, we can track whether a specific relationship encoded in the input embedding is remembered by the RNN. It has been shown that relationships between word embeddings can be characterized as vector offsets, e.g. the male -- female and common person -- royal person relationships, as demonstrated by the famous example `king - man + woman = queen'~\cite{Mik:13}. 
In order to measure to which extent linear relationships in embedding space are retained in hidden state space, we again make use of the SVD of the state gradients, assuming that some linear relationships in state space are present. The directions in the embedding space with the largest SVs are best preserved in the hidden state space. Hence, we can compare a vector characterizing a specific property to those directions to see how well the property is remembered in the hidden state.

In what follows, we will first  give an overview of related work (section~\ref{related}). Then we explain how our approach works in section~\ref{method} and demonstrate it with experimental results in section~\ref{exp}. We end with conclusions and an outlook to future work (section~\ref{concl}).

\section{Related work}
\label{related}


A general framework to explain the predictions of a classifier is presented in~\cite{lime}, but it can only work with interpretable data representations such as binary vectors, whereas we present a framework that can deal with continuous (word) embeddings.

In vision, several backpropagation-based techniques for visualization and investigation of the inner workings of neural networks have been proposed~\cite{Erhan,Simonyan}. These methods differ from our approach in the sense that they compute gradients of the output of the network with respect to the input, like in classical backpropagation.



Hermans and Schrauwen~\cite{Hermans} examine the influence of each layer in a character-level LM by setting its output to 0 and calculate the distance between hidden states after processing sequences that are identical except for one typo.
Li et al.~\cite{JiweiLi} use several techniques for visualization:
t-SNE visualization~\cite{t-SNE}, a heatmap of the neuron activations over time, the first-derivative saliency method~\cite{Erhan,Simonyan}, and the variance of a specific word embedding with respect to an averaged word embedding. Ji et al.~\cite{JiweiLi-b} make use of representation erasure: measuring the contribution of individual input units by erasing them and seeing how it affects the performance of the model. Strobelt et al.~\cite{Strobelt} release a tool that visualizes the evolution in LSTM hidden states for several text processing tasks.
Karpathy et al.~\cite{Karpathy} visualize the activations of individual cells,
plot gate activation statistics and perform error analysis. In this paper, we do not visualize the hidden states themselves, but rather the influence of the input on the hidden states.

Another approach is trying to understand the reason why the network makes a certain prediction. Lei et al.~\cite{Lei} extract the parts of the input that are important for predicting the output, while Alvarez-Melis and Jaakkola~\cite{Alvarez} provide a general framework that can explain the predictions of models with structured input and output. 

Analyses can also be used to improve models, as Aubakirova and Bansal~\cite{Aubakirova} demonstrate:~they look at the activations of CNN models trained for politeness prediction and discover new features that can improve feature-based models. They also use the first-derivative saliency method to show how much each input contributes to the final decision of the model. Adi et al.~\cite{Adi} analyze sentence embeddings by using them as input for a classifier that is trained to predict sentence length, word content and word order. 

Finally, work has also been done towards examining the ability of NNs to model linguistic phenomena such as subject -- verb agreement~\cite{Linzen,Bernardy}. Broere~\cite{Broere} investigates the syntactic properties of skip-thought sentence representations by training logistic regression on them to predict POS tags and dependency relations. In a similar vein, we predict POS tags from word embeddings to verify whether specific classes are linearly separable in the embedding space.




\section{SVD of state gradients}
\label{method}

\subsection{Average memory of the RNN}

We calculate the gradients of every component in the state of the RNN with respect to every component in the input embedding of time step $t - \tau$, where $t$ is the current time step and $\tau$ the delay. For example, if the input sentence is \textit{the cat lies on the mat} and we are currently processing the word \textit{mat}, the input for a delay of 0 is \textit{mat} itself, the input for a delay of 1 is \textit{the} etc. For a specific time step and specific delay, our gradients can be arranged in a gradient matrix $\mathbf{G}_{t,\tau}$ (size of the embedding $\times$ size of the hidden state). Averaging over all time steps for a certain delay $\tau$ gives us the average gradient matrix for $\tau$: $\bar{\mathbf{G}}_\tau$. This matrix hence contains the average influence of every component of the input embedding $\tau$ steps later on every component of the RNN state.

We decompose the average gradient matrix with SVD:

\begin{equation}
\bar{\mathbf{G}}_\tau = \mathbf{U}~\mathbf{\Sigma}~\mathbf{V}^T = \sigma_1~\mathbf{u}_1~\mathbf{v}_1^T + \sigma_2~\mathbf{u}_2~\mathbf{v}_2^T + \ldots
\end{equation}
in which $\mathbf{U}$ and $\mathbf{V}$ are orthogonal matrices (with columns $\mathbf{u}_i$ and $\mathbf{v}_i$) and $\mathbf{\Sigma}$ is a rectangular diagonal matrix with the singular values $\sigma_i$. We can interpret $\mathbf{V}$ as directions in the embedding space, $\mathbf{\Sigma}$ as the extent to which the directions in the embedding space can be found in the hidden state space and $\mathbf{U}$ as corresponding directions in the hidden state space. Hence, the directions with the largest SVs (lowest index) are directions in embedding space that are best remembered by the RNN. 

In order to investigate how well the RNN remembers on a corpus level, we can track the largest SV or the sum of all SVs with respect to the delay $\tau$. We can also compare the SVs based on gradient matrices averaged over specific classes of words only (see section~\ref{exp-corpus}), or even over the occurrences of individual words. 

\subsection{Tracking a specific property}
\label{property}

If we want to know to what extent a specific property encoded in the input embedding is remembered in the state, we will compare the vector encoding this property to the directions in $\mathbf{V}^T$ corresponding to the largest SVs of the average gradient matrix.

We calculate the extent to which a certain property, e.g. the difference between singular (`sg') and plural (`pl'), is remembered as follows: we first define a property as the difference between the averaged embeddings for the classes separated by that property.

\begin{equation}
\mathbf{d}_{a-b} = \bar{\mathbf{e}}_a - \bar{\mathbf{e}}_b
\end{equation}
where $\bar{\mathbf{e}}_a$ and $\bar{\mathbf{e}}_b$ are the result of averaging all embeddings of words belonging to classes $a$ and $b$ respectively. Before defining a specific property as the difference between the average embeddings, we will first check whether the embeddings of the two classes are linearly separable by training a linear classifier (see section~\ref{pos-classification}). 

We propose two methods to investigate the extent to which a property is remembered. Firstly, we can compare $\mathbf{d}$ with the subspace of the embedding space spanned by the directions that are best remembered, the $n$ largest right-singular vectors, $\mathcal{H}_n$. To be able to do this, we make the orthogonal projection of $\mathbf{d}$ on $\mathcal{H}_n$:

\begin{equation}
\mathbf{y} = \proj_{\mathcal{H}_n}~\mathbf{d} = \mathbf{V}_n~\mathbf{V}_n^T~\mathbf{d}
\end{equation}
where $\mathbf{V}_n$ is the matrix containing the $n$ first columns of $\mathbf{V}$. Assuming $\mathbf{d}$ is normalized to unit length, we can calculate the cosine similarity between $\mathbf{y}$ and $\mathbf{d}$ as follows:

\begin{equation}
\cos(\mathbf{d},\mathcal{H}_n) = \cos(\mathbf{d},\mathbf{y}) = \frac{\mathbf{d}^T~\mathbf{V}_n~\mathbf{V}_n^T~\mathbf{d}}{\left\lVert \mathbf{d} \right\rVert~\left\lVert \mathbf{V}_n~\mathbf{V}_n^T~\mathbf{d} \right\rVert} = \left\lVert \mathbf{V}_n^T~\mathbf{d} \right\rVert
\end{equation}
The cosine similarity between $\mathbf{d}$ and $\mathcal{H}_n$ is a measure of how close $\mathbf{d}$ is to the top $n$ directions that are best remembered in the RNN state: the closer to 1, the better the property specified by $\mathbf{d}$ is remembered.

A second option is comparing $\mathbf{d}$ with the direction in embedding space that is \textit{best} remembered. To do this, we multiply $\mathbf{d}$ with the average gradient matrix:


\begin{equation}
\label{eq:r}
r = \left\lVert \bar{\mathbf{G}}_\tau \times \mathbf{d} \right\rVert
\end{equation}
where $r$ is a measure of the extent to which the difference between classes $a$ and $b$ is remembered by the state. If $\mathbf{d}$ would be the embedding direction that is best remembered in the state, then it would be equal to $\mathbf{v}_1$ and the result of the multiplication in equation~\ref{eq:r} would be $\sigma_1$.
Hence, in order to get a relative measure of how well the difference between two classes is remembered, we compare $r$ with $\sigma_1$ and obtain a `extent to which the property is remembered, relative to the property that is best remembered', which we will henceforth refer to as the `relative memory' $m$:

\begin{equation}
m = \frac{r}{\sigma_1}
\end{equation}

Notice that the cosine similarity between $\mathbf{d}$ and $\mathcal{H}_n$ would be equal to the relative memory if $\sigma_n = \sigma_1$ and $\sigma_{n+1} = 0$. Thus, both measures capture the extent to which a certain property is remembered, but the cosine similarity only compares $\mathbf{d}$ with the first $n$ right-singular vectors while taking into account the strength with which those directions are remembered, whereas the relative memory compares with all right-singular vectors but is a measure relative to the direction that is best remembered. In section~\ref{exp-property}, we present results for both measures.

\section{Experiments}
\label{exp}

\subsection{Setup}

We will focus on the cell state or memory of the LSTM as state, but note that similar experiments can be done for the hidden state/output of the LSTM or states of other RNNs.





We train LSTM LMs on the widely used Penn TreeBank (PTB) benchmark~\cite{ptb}, that contains 900k word tokens for training, 70k word tokens as validation set and 80k words as test set. We chose PTB because it contains manually assigned part-of-speech (POS) tags that we will use to train linear classifiers, but 900k words is quite small. Thus, we also train embeddings on Wall Street Journal, which encompasses PTB and is hence in-domain data, but is much larger: we use the CSR LM-1 corpus (LDC) with non-verbalized punctuation (years 87--94) which contains 110M words. 
Embeddings trained on these data give better classification results than embeddings trained on PTB only (see section~\ref{pos-classification}) and are hence better linearly separable. Thus, we want to use pre-trained WSJ embeddings in our PTB LMs too. The LM with these embeddings is henceforth referred to as `cbow WSJ', whereas the LM with embeddings trained from scratch is called `joint.tr.'. However, since the version of PTB that is commonly used for language modeling~\cite{Mik:10} is normalized differently than WSJ, the vocabularies of the two data sets do not match. If we want to use pre-trained WSJ embeddings, we need an embedding for every word in the PTB corpus.
Hence, we chose to do some additional normalization on PTB (henceforth referred to as `PTB-norm'), e.g.~removing hyphens from certain words (e.g. `company-owned' $\rightarrow$ `company owned').
We only changed one thing for the WSJ data (`WSJ-norm'): since numbers are converted to `N' in PTB and inverting this operation is much more difficult, we chose to convert numbers in WSJ to `N' instead. The resulting vocabulary for PTB-norm contains 10921 words instead of the usual 10k words.

We use TensorFlow~\cite{tensorflow} to train the LSTMs and to compute the gradient matrices. The linear classifiers are trained with scikit-learn~\cite{sklearn}, with a grid search over several hyperparameters: regularization (L1, L2), regularization strength, initialization seed, frequency-based class weights or not, one vs all or multinomial loss and optimization algorithm. 
The word2vec embeddings (same size as the LM embeddings) are trained with the default word2vec parameters (cbow).

The LSTM LM consists of an embedding layer of dimension 64 and 1 layer of 256 LSTM cells. Using a larger embedding size in combination with dropout~\cite{dropout} on the embeddings gives slightly better results, but we choose to not use dropout on the embeddings/input of the LSTM cell since we are interested in seeing what the model learns from the input, and masking the input might have unexpected effects on our analysis. We do apply dropout on the output of the LSTM cells with a probability of 50\%. 
The norm of the gradients is clipped at 5. We train on batches of size 20, each batch element containing a text sequence of 50 words. 
The LSTM is trained with stochastic gradient descent, starting with a learning rate of 1 for the first 6 epochs, after which it is exponentially decreased with a decay of 0.8. As a reference, this model has a validation perplexity of 81.5 and a test perplexity of 78.8. If we use cbow embeddings trained on WSJ to train an LM (without re-training the embeddings), they perform worse in the LM: perplexities 87.0 (validation) and 83.0 (test), even though they give better logistic regression accuracies. 

\subsection{Average memory}
\label{exp-corpus}

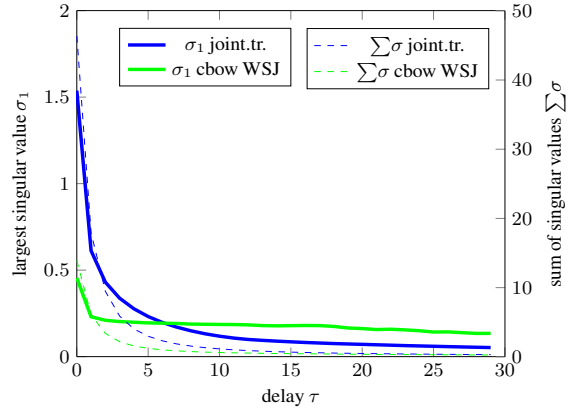
\begin{figure}[t]
\centering
\resizebox{7.5cm}{5.5cm}{\input{largest_sv_delay.tex}}
\setlength{\abovecaptionskip}{0cm}
\setlength{\belowcaptionskip}{-10pt}
\caption{Largest SV and sum of SVs of the average gradient matrix with respect to delay between the cell state and the input, calculated on the validation set of PTB-norm.}
\label{fig:largest_sv_delay}
\end{figure}

In Figure~\ref{fig:largest_sv_delay}, we plot the largest SV $\sigma_1$ and the sum of all SVs $\sum{\sigma}$ for the average gradient matrix per delay. 
For the LM with jointly trained embeddings (blue), there is a sharp decrease in the first part of the plot, indicating that much of the information that is present in the cell state about the current word ($\tau = 0$) is quickly forgotten after about 5 words. However, on average, some information is still remembered even after processing more than 20 words. If we compare the trends of $\sigma_1$ and $\sum{\sigma}$ (notice the different scales on the left and right y-axes), we see that the largest SV decreases relatively slower, which seems to indicate that the memory becomes more selective when the delay increases: the ratio $\sigma_1 / \sum{\sigma}$ becomes larger. 

The plots for the LM with pre-trained WSJ embeddings (green) show not only smaller absolute values but also a less sharp decrease. This model is much more selective when the delay increases: for a delay of 29, the ratio $\sigma_1 / \sum{\sigma}$ is 50\% compared to 16\% for the first LM. A possible explanation for this is that the cbow embeddings (inherently limited to the short term) do not contain certain information that is important for the LM on the long term and hence the LM can remember less relevant information.


If we compare $\sigma_1$ for different classes of input words (see column `LM joint.tr.' in Table~\ref{tab:sv-classes}), we notice some tendencies that intuitively make sense. 
We observe that pronouns have the largest effect on the cell state, followed by nouns. This makes sense since these word classes are important from a syntactical (e.g. pronouns/nouns having an effect on which verb conjugation should follow) and/or semantic (nouns carrying much of the meaning of the sentence) point of view. In a similar vein, verbs can also carry semantics and are important in predicting the syntactic role or POS from the next word(s) (e.g. a specific preposition, a semantic class $\ldots$). On the other hand, adjectives, adverbs, conjunctions and prepositions are less informative in predicting the next words. 
In the last column of Table~\ref{tab:sv-classes}, we present the largest SVs for the LM with cbow WSJ embeddings. We only compare the relative relationships between the POS classes, since the absolute values of the SVs are based on different vector spaces. Even though the class that is best remembered, pronouns, stays the same, we see some tendencies that are counterintuitive. Adverbs, adjectives and conjunctions/prepositions are better remembered than nouns and verbs. We also observe that the variance between the POS classes is larger for this LM, as the normalized values (between brackets) indicate. Clearly there is quite some difference in the manner in which the two LMs process the embeddings, which is an interesting topic for further investigation.



%
%

  

\begin{table}[t]
  \caption{Largest SV $\sigma_1$ ($\tau$ = 0) per POS. The values between brackets are normalized SVs.}
  \label{tab:sv-classes}
  \centering
  \begin{tabular}{ l  c  c}
    \toprule
    \multicolumn{1}{l}{\textbf{Class}} &  \multicolumn{1}{c}{\textbf{joint.tr.}} & 
    			\multicolumn{1}{c}{\textbf{cbow WSJ}}\\
    \midrule
    pronouns				& $1.95$ $(0.20)$ 	& $0.96$ $(0.27)$				\\
	nouns					& $1.66$ $(0.17)$ 	& $0.42$ $(0.12)$				\\
    verbs					& $1.65$ $(0.17)$ 	& $0.48$ $(0.13)$				\\
    adjectives				& $1.57$ $(0.16)$ 	& $0.54$ $(0.15)$				\\
    adverbs					& $1.44$ $(0.15)$ 	& $0.59$ $(0.17)$				\\
    conj./prep.				& $1.44$ $(0.15)$	& $0.54$ $(0.15)$ 				\\
    \bottomrule
  \end{tabular}
\vspace{-1.0em}
\end{table}

\subsection{Linear classification}
\label{pos-classification}

Before investigating the extent to which a difference vector for a certain property is remembered in the cell state, we first check whether it makes sense to categorize a certain property as a difference vector by verifying whether the two classes are separable by a linear classifier.

\begin{table}[t]
  \caption{Validation accuracy of logistic regression to predict POS-based classes. Numbers between brackets = number of target classes. N=nouns, V=verbs, Adj=adjectives, Adv=adverbs, Pro=pronouns.}
  \label{tab:linear-class}
  \centering
  \begin{tabular}{ l c c c c}
    \toprule
    \multicolumn{1}{l}{\textbf{}} & \multicolumn{2}{c}{\textbf{PTB-norm}}
    	& \multicolumn{2}{c}{\textbf{WSJ-norm}} \\
    \multicolumn{1}{l}{\textbf{Class}} &	\multicolumn{1}{c}{\textbf{cbow}} & 
    	\multicolumn{1}{c}{\textbf{LM emb}} &  \multicolumn{1}{c}{\textbf{cbow}} &
        \multicolumn{1}{c}{\textbf{LM emb}}\\
    \midrule
    all (34)				& $35.99$	& $42.39$ 	& $\mathbf{59.44}$	& $58.59$		\\
	nouns (4)				& $71.23$	& $73.54$ 	& $89.69$	& $\mathbf{90.15}$		\\
    \hline
    N-V-Adj-Adv-Pro 			& $57.34$	& $66.33$ 	& $75.00$	& $\mathbf{75.04}$	\\
    N-V			   			& $73.60$	& $77.28$ 	& $\mathbf{84.65}$	& $83.42$		\\
    N-Adj			   		& $79.06$	& $84.47$ 	& $87.53$	& $\mathbf{88.59}$		\\
    V-Adv			   		& $91.48$	& $95.00$ 	& $95.93$	& $\mathbf{97.78}$		\\
    \hline
    sg-pl 					& $84.31$	& $87.08$ 	& $95.85$	& $\mathbf{96.31}$		\\
    common-proper 			& $84.46$	& $83.69$ 	& $91.54$	& $\mathbf{92.15}$		\\
    \bottomrule
  \end{tabular}
\vspace{-1.0em}
\end{table}

In Table~\ref{tab:linear-class}, we present the validation accuracies for different settings of the classifier and different types of word representations. If we compare the type of word representations in the first two columns, we see that in most cases, the embeddings that are jointly trained with the LM (`LM emb') are better than the cbow embeddings in this task. 
Training the embeddings on more, in-domain, data (WSJ) also increases the classification accuracy. Using the embeddings of an LM trained on WSJ usually gives slightly better results than cbow embeddings, but given that the perplexity of the PTB LM with those embeddings is significantly higher (107.2 compared to 83.0 for cbow embeddings), we choose to focus on the cbow WSJ embeddings.

The first part of the table contains results for fine-grained distinctions. Trying to predict all POS tags jointly (first row) is clearly too difficult, probably because the number of classes is much larger (34) and because there are quite some infrequent classes.  Distinguishing between the different sub-types for a class, e.g. sg, pl, sg proper and pl proper for the class of nouns, is easier. 
The second part of the table contains results for more coarse-grained distinctions, e.g. `N-V' is a classifier that tries to predict whether the embeddings belong to a noun or a verb. 
In general, we observe that the classifier achieves reasonable accuracies if the number of classes to separate is limited.

The final part of the table are examples of specific properties that can be derived from the PTB POS tags. We observe that both the distinction sg -- pl noun and common -- proper noun gives a reasonable accuracy. Hence, we will investigate how those properties are remembered in the LSTM cell in the next section.

\subsection{Tracking specific properties}
\label{exp-property}

\begin{figure}[t]
\resizebox{8.4cm}{6.7cm}{\input{ratio+cos_sg-pl_common-proper.tex}}
\caption{$m$ and $\cos(\mathbf{d},\mathcal{H}_5)$ for sg -- pl and common -- proper nouns.}
\label{fig:ratio+cos_sg-pl_common-proper}
\vspace{-1.0em}
\end{figure}
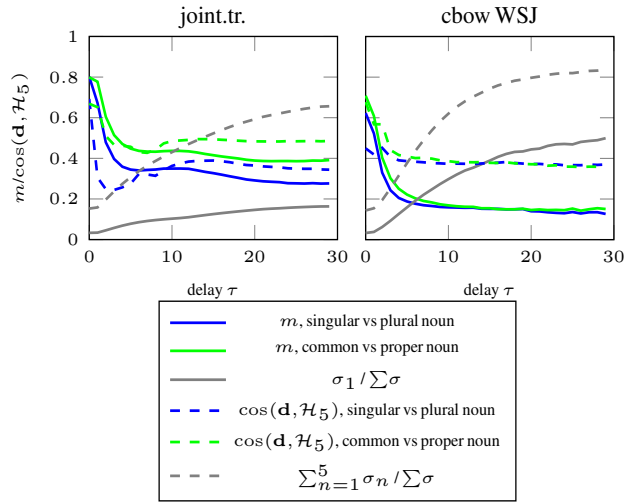

In Figure~\ref{fig:ratio+cos_sg-pl_common-proper}, we plot for the difference vectors separating singular versus plural nouns and common versus proper nouns the relative memory $m$ and the cosine similarities with $\mathcal{H}_{5}$. The cosine similarities between the two difference vectors is -0.59 for the first LM (left) and -0.57 for the second LM (right), so there is a clear distinction between them.

For the LM with jointly trained embeddings, we see that according to both measures the sg -- pl distinction is slightly better remembered for a delay of 0, while for the other delays the common -- proper distinction is better remembered. There is a dip for the cosine similarities between delays 1 -- 9 (sg -- pl) and 2 -- 8 (common -- proper) and a sharp decrease for the relative memory at a delay of 2, indicating that the properties seem mostly important on the short term. However, the distinction stays in memory even for longer delays, as both measures show.
We also plot the ratio of $\sigma_1$ and the sum of the 5 largest SVs with respect to the sum of all SVs (gray lines). Notice that if the delay increases, the ratio increases too, which confirms our observation for Figure~\ref{fig:largest_sv_delay} that the memory becomes more selective over time.

For the LM with pre-trained WSJ embeddings, we see similar trends as for the other LM, namely that the discussed properties are mostly important on the short term and that generally, the distinction common -- proper is better remembered than sg -- pl. However, the difference between the properties is much smaller and even more so for longer delays. Since for delays larger than 10, $m$ is only about 0.2  and the cosine similarities are around 0.4 even though we are comparing the difference vectors with about 60\% (for delays between 10 and 20) to 80\% (for delays larger than 20) of the original embedding space, this seems to suggest that this model does not remember these properties very well on the long term, which is in line with our previous observation that this LM is very selective on the long term (see section~\ref{exp-corpus}).

\section{Conclusion and future work}
\label{concl}

We present a framework to analyze what the states of an RNN remember from their input and for how long. The approach is based on the gradients of the states with respect to the input word embeddings with a certain delay. We apply this framework to the cell state of LSTM language models. Based on SVD on the gradient matrix we inspect how long the RNN remembers on average or for specific word classes. We can also track to what extent a specific property is remembered by comparing it with directions in embedding space that are best transferred to the hidden state space.

We observe that the LSTM LM is capable of remembering at least part of the input from 30 time steps earlier, but becomes much more selective when the delay increases. Using embeddings pre-trained on a much larger, in-domain dataset not only increased the perplexity of the model but also made its memory more selective, indicating that the pre-trained embedding lack certain (long-range) relevant information. Additionally, we show that the LSTM LM remembers certain word classes such as pronouns more strongly than others and that the property of singular versus plural is slightly better remembered than common versus proper noun.

In the future, we would like to extend this research by 
tracking different properties, doing a more extensive comparison (more values of $n$, larger delays) and looking at more specific contexts.
It would also be interesting to compare with the LSTM hidden state/output and with other types of RNNs, and to examine the influence of certain hyperparameters, such as the size of the hidden layer.


\section{Acknowledgements}

This research is funded by the Flemish government agency IWT (project 130041, SCATE). 

\bibliographystyle{IEEEtran}

\bibliography{mybib}


\end{document}

%% file: largest_sv_delay.tex
\begin{tikzpicture}
\begin{axis}[
    xmin=0, xmax=30,
    ymin=0, ymax=2,
	xlabel={delay $\tau$},
    axis y line*=left,
	ylabel={largest singular value $\sigma_1$},
    every axis plot/.append style={ultra thick},
    legend style={at={(0.3,0.95)},anchor=north}
]
\addplot[blue,mark=circle*] table {largest_sv.dat};
\addlegendentry{$\sigma_1$ joint.tr.};
\addplot[green,mark=circle*] table {pretr_largest_sv.dat};
\addlegendentry{$\sigma_1$ cbow WSJ};

\end{axis}
\begin{axis}[
    xmin=0, xmax=30,
    ymin=0, ymax=50,
    hide x axis,
    ylabel near ticks,
    axis y line*=right,
	ylabel={sum of singular values $\sum{\sigma}$},
    legend style={at={(0.75,0.95)},anchor=north}
]
\addplot[blue,dashed,mark=circle*] table {sum_svs.dat};
\addlegendentry{$\sum{\sigma}$ joint.tr.};
\addplot[green,dashed,mark=circle*] table {pretr_sum_svs.dat};
\addlegendentry{$\sum{\sigma}$ cbow WSJ};

\end{axis}
\end{tikzpicture}

%% file: ratio+cos_sg-pl_common-proper.tex
\begin{tikzpicture}
\begin{groupplot}[
	group style={group size=2 by 3,
    horizontal sep=0.3cm},
    width=0.25\textwidth]
\nextgroupplot[
	title={joint.tr.},
    title style={at={(0.5,1.1)},anchor=north,font=\scriptsize},
    xmin=0, xmax=30,
    ymin=0, ymax=1,
	xticklabel style = {font=\tiny},
    yticklabel style = {font=\tiny},
	xlabel={delay $\tau$},
    ylabel={$m$/$\cos(\mathbf{d},\mathcal{H}_5)$},
    label style={font=\tiny},
    every axis plot/.append style={thick},
    legend style={at={(1,-0.3)},anchor=north,font=\tiny}
]
\addplot[blue,mark=circle*] table {1sg-pl.dat};
\addlegendentry{$m$, singular vs plural noun};
\addplot[green,mark=circle*] table {1common-proper.dat};
\addlegendentry{$m$, common vs proper noun};
\addplot[gray,mark=circle*] table {perc_total_sum_svs_n1.dat};
\addlegendentry{$\sigma_1$ / $\sum{\sigma}$};
\addplot[blue,dashed,mark=circle*] table {cos_sg-pl_n5.dat};
\addlegendentry{$\cos(\mathbf{d},\mathcal{H}_5)$, singular vs plural noun};
\addplot[green,dashed,mark=circle*] table {cos_common-proper_n5.dat};
\addlegendentry{$\cos(\mathbf{d},\mathcal{H}_5)$, common vs proper noun};
\addplot[gray,dashed,mark=circle*] table {perc_total_sum_svs_n5.dat};
\addlegendentry{$\sum_{n=1}^5{\sigma_n}$ / $\sum{\sigma}$};

\nextgroupplot[
	title={cbow WSJ},
    title style={at={(0.5,1.1)},anchor=north,font=\scriptsize},
    xmin=0, xmax=30,
    ymin=0, ymax=1,
	xticklabel style = {font=\tiny},
    yticklabels=\empty,
	xlabel={delay $\tau$},
    label style={font=\tiny},
    every axis plot/.append style={thick}
]
\addplot[blue,mark=circle*] table {1sg-pl_pretr.dat};
\addplot[green,mark=circle*] table {1common-proper_pretr.dat};
\addplot[gray,mark=circle*] table {pretr_perc_total_sum_svs_n1.dat};
\addplot[blue,dashed,mark=circle*] table {pretr_cos_sg-pl_n5.dat};
\addplot[green,dashed,mark=circle*] table {pretr_cos_common-proper_n5.dat};
\addplot[gray,dashed,mark=circle*] table {pretr_perc_total_sum_svs_n5.dat};

\end{groupplot}
\end{tikzpicture}